\documentclass[11pt]{article}

\usepackage{acl}

\usepackage{times}
\usepackage{latexsym}
\usepackage{tabularx}
\usepackage{graphicx}
\usepackage{hyperref}
\usepackage{enumerate}
\usepackage{url}
\usepackage{multirow}
\usepackage{geometry} 
\usepackage[linesnumbered,ruled,vlined]{algorithm2e}
\usepackage{algpseudocode}
\usepackage{amsmath}
\usepackage{arydshln} 
\usepackage{comment}
\usepackage{subcaption}
\usepackage{float}
\usepackage{adjustbox}

\usepackage[T1]{fontenc}
\usepackage[utf8]{inputenc}
\usepackage{microtype}
\usepackage{inconsolata}

\urldef{\zenodoURL}\url{https://zenodo.org/record/7152317#.Yz6mJ9JByC0}
\usepackage[T1]{fontenc}

\usepackage[utf8]{inputenc}

\usepackage{microtype}

\usepackage{inconsolata}

\usepackage{lineno}

\title{Citation-Based Summarization of Landmark Judgments}

\author{Purnima Bindal \\
  University of Delhi, India \\
 \texttt{pbindal@cs.du.ac.in} \And
  Vikas Kumar \\
  University of Delhi, India \\
 \texttt{vikas@cs.du.ac.in}  \And
  Vasudha Bhatnagar \\
  University of Delhi, India \\
 \texttt{vbhatnagar@cs.du.ac.in} \AND
  Parikshet Sirohi \\
  University of Delhi, India \\
 \texttt{parikshet.sirohi@gmail.com}  \And
  Ashwini Siwal  \\
  University of Delhi, India \\
 \texttt{asiwal@law.du.ac.in} 
 \\}

\begin{document}
\maketitle
\begin{abstract}
Landmark judgments are of prime importance in the Common Law System because of their exceptional jurisprudence and frequent references in other judgments.  In this work, we leverage contextual references available in citing judgments to create an extractive summary of the target judgment. We evaluate the proposed algorithm on two datasets curated from the judgments of Indian Courts and find the results promising.

\end{abstract}
%
%
%

\section{Introduction}
\label{sec-introduction}
In a  Common law legal framework, law professionals scrutinize a plethora of legal case documents to comprehend the Court’s handling in diverse legal scenarios. These documents customarily span from dozens to hundreds of pages, making them arduous to comprehend. Ergo,  condensed summaries are a valuable aid for legal research. Manually crafting case summaries is an intellectually demanding task that requires deep, intense legal knowledge and experience of law experts. Automatic summarization of legal judgments is a practical and potent solution and, therefore, a widely researched problem among NLP researchers. Several earlier studies have provided insights into the challenges associated with summarization of judgments~\citeyearpar{farzindar2004atefeh, saravanan2006improving, bhattacharya2019comparative, parikh2021lawsum, feijo2023improving}. Summarizing Indian judgments is tough due to their varied structures, unlike US, UK, Australia and Canada.

Landmark judgments are important court rulings that establish novel legal principles, handle remarkable legal issues, mould the understanding of law, and leave a long-lasting effect on jurisprudence. These judgments not only draw public attention but also gather a large number of citations in later judgments. The citing judgments spotlight the substantive arguments and precedents that lend weight to the ruling in the cited judgment. Each citation in a citing judgment is an informative source for legal points of the prior case, statutes, or laws to support the decision. Summary of a \textit{landmark} judgment highlights the noteworthy attributes of the judgment and provides insight into issues handled within the case, critical legal points, and arguments presented by the lawyers. \\

\noindent\textit{Related Work}:
 Legal text summarization has drawn considerable attention, specifically in the Indian context
\citeyearpar{saravanan2006improving,bhattacharya2019comparative,jain2021summarization,furniturewala2021legal,chalkidis2021lexglue,bhattacharya2021incorporating,Shukla2022,kalamkar2022named, ghosh2022indian, anand2022effective}. \citeauthor{bhattacharya2019comparative} compare legal summarization algorithms on Indian Supreme Court judgments dataset\footnote{\zenodoURL}, while \citet{Shukla2022} comprehensively evaluate extractive and abstractive summarization algorithms through automated and human assessment. Extractive legal-domain specific summarization algorithms in other countries  include \citet{farzindar2004atefeh} (Canada),    \citet{galgani2012citation, polsley2016casesummarizer,galgani2012combining} (Australia),  \citet{liu2019extracting} (China), \citet{bhattacharya2021incorporating} (India). %

Abstractive Legal Text Summarization  algoritrhms include   Legal-LED \cite{nsi319_legal_led_base_16384}, Legal-pegasus \cite{nsi319/legal-pegasus}, LegalSumm \citep{feijo2023improving}. Recently,  \citet{paul2022pre} developed two transformer-based pre-trained language models, InLegalBERT and InCaseLawBERT, through the re-training of  LegalBERT \citep{chalkidis2021lexglue} and CaseLawBERT \citep{zheng2021does}, respectively.

\noindent\textit{Contributions:}  We design an unsupervised,  extractive algorithm, \textit{CB-JSumm}, to summarize a landmark judgment by leveraging its incoming citations (Sec. \ref{sec: method}).  We also curate two datasets for citation-based summarization of judgments consisting of  $99$ ($50 + 49$) Indian judgments, citing judgments, and gold standard (reference) summaries and assess the quality of the algorithmic summaries(Sec. \ref{sec:Experimental Results}).


\section{Methodology}
\label{sec: method}

Given a target landmark judgment $\mathbf{J}$ to be summarized, and $\mathbf{C} = \{\mathbf{C}_{1}, \ldots, \mathbf{C}_{t}\}$ a set of \textit{citing judgments}, the goal of citation-based legal text summarizer is to present a condensed representation of $\mathbf{J}$ using the \textit{citation }information available in the citing judgments ($\mathbf{C}_i$'s). Judgment $\mathbf{J}$ is tokenized into $n$ sentences $(\mathbf{j_1}, \ldots, \mathbf{j_n})$ and salient sentences are selected for summary $JSumm$. 

 For citation-based summarization, it is prudent to consider the context in which the judgment is referred. However, extracting context from the citing judgment is a nontrivial task and requires a cautious approach. Since the referring sentence may be inadequate representative of the context in which the target judgment $\mathbf{J}$ is referred, we consider the entire paragraph where the target judgment is cited. This paragraph is referred as the \textit{citing-text-span} and individual sentences in \textit{citing-text-span} are referred as \textit{citation sentences} or simply \textit{citances}.

\begin{figure}
    \centering
    \adjustbox{max width=\linewidth}{    
     \includegraphics[width=\textwidth]{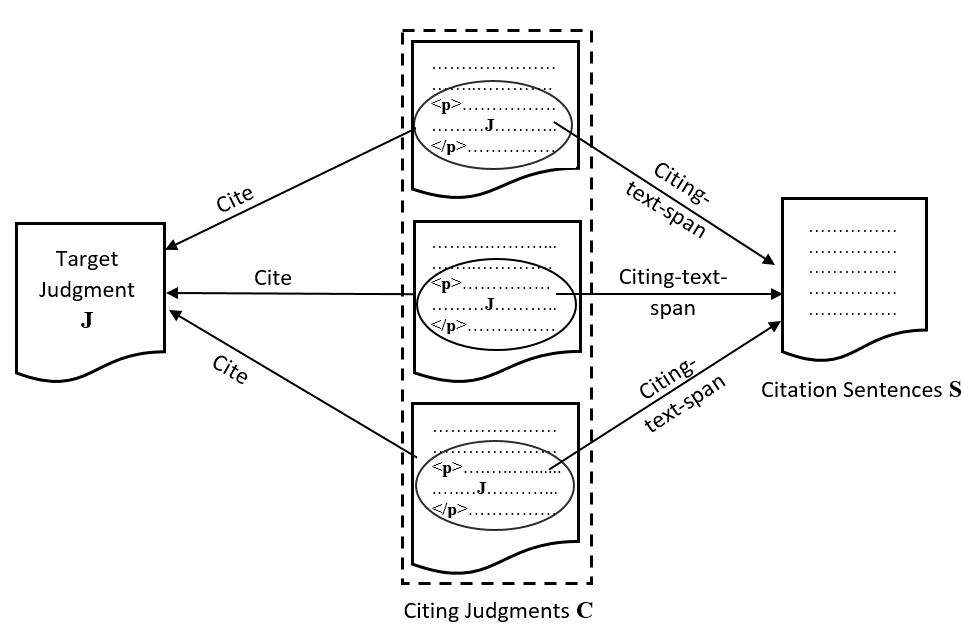} 
     }
    \caption{A target landmark judgment $\mathbf J$, set $\mathbf C$ of citing judgments, and the corpus of extracted citation sentences $\mathbf S$ }
    \label{fig:citingJudgmets}
\end{figure}

All \textit{citing-text-spans} are harvested from all citing judgments of $\mathbf{J}$ and tokenized into sentences. The collection of citances from  the  \textit{citing-text-spans} is denoted by  $\mathbf{S}$. Thus, $\mathbf{S} = (\mathbf{s_1}, \ldots, \mathbf{s_m} )$ contains all contextual information from all citing judgments contained in $\mathbf{C}$. Fig. \ref{fig:citingJudgmets} clarifies the notation and terminology used throughout the paper. 

\subsection{Citation Based Judgment Summarization Algorithm}
  The core idea of the proposed algorithm \textit{CB-JSumm} (\textit{C}itation-\textit{B}ased \textit{J}udgment \textit{Summ}arization) is to leverage contextual information contained in the \textit{citing-text-spans} of the citing judgments for identifying the significant sentences in the target judgment. The algorithm, which has three phases,  requires citing judgments of the target judgment ($\mathbf{J}$) to prepare the input corpus of citances ($\mathbf{S}$). 
  
\begin{figure}
    \centering
    \adjustbox{max width=\linewidth}{
    \includegraphics[width=0.5\textwidth]{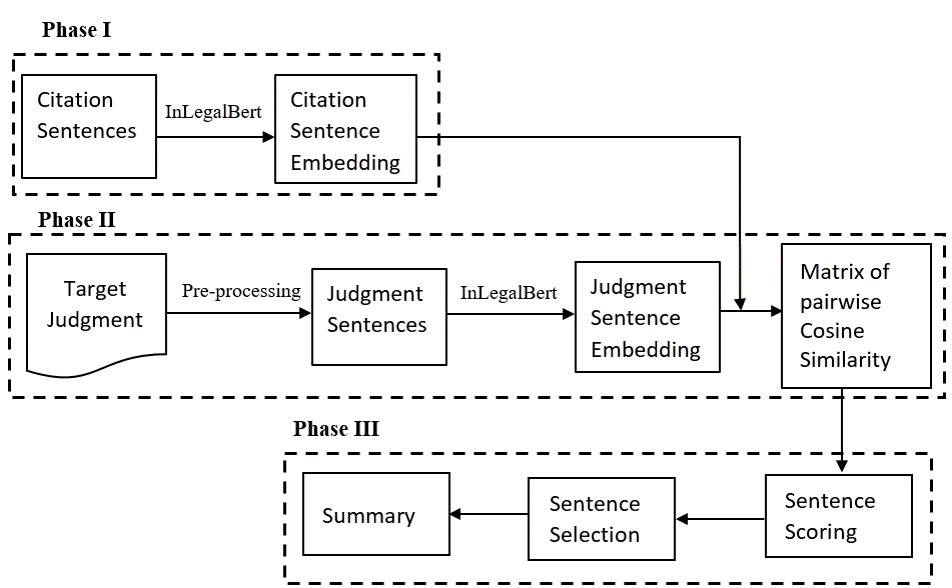}
    }
    \caption{Pipeline for proposed \textit{CB-JSumm} Algorithm}
    \label{fig:algorithm-pipeline}
\end{figure}

  Phase I of the algorithm is the preparatory step to retrieve contextual embeddings of sentences in $\mathbf{S}$. For
 this purpose, we use InLegalBert, a pre-trained transformer-based language model tailored for the Indian legal domain \cite{paul2022pre}. Similarly, we retrieve embeddings for the sentences in $\mathbf{J}$ and compute semantic similarity between the citances and judgment sentences in phase II (Fig. \ref{fig:algorithm-pipeline}). Finally, we identify judgment sentences that are worthy of being included in the summary based on the semantic similarity scores in phase III. 
\begin{algorithm}
	\scriptsize
        \KwIn{ set of $n$ judgment sentences $\mathbf{J}$, set of $m$ citation sentences $\mathbf{S}$, desired summary length $\mathit {l}$}
        \KwOut{Judgment Summary $JSumm$ }
        $S_E$ $\leftarrow$ Embedding($\mathbf{S}$); // Embeddings  of $m$ citation sentences \\
        $J_E$ $\leftarrow$ Embedding($\mathbf{J}$); // Embeddings  of $n$ judgment sentences\\
        $\mathcal S_{m \times n}$  $\leftarrow$ cosine\mbox{-}sim($S_E, J_E)$; // Computing similarity score\\
        $JSumm$ $\leftarrow$ sentence\mbox{-}scoring($\mathcal S, \mathit {l})$; // using Algorithms in Sec. \ref{subSec:Sentence Scoring} \\ 
 \caption{\textit{CB-JSumm} Algorithm}
\label{algo: CB-JSumm}
\end{algorithm}

Algorithm \ref{algo: CB-JSumm} outlines the proposed \textit{CB-JSumm} algorithm. In Steps 1 and 2, we retrieve contextual embeddings of $m$  citances in $\mathbf S$  and $n$ judgment sentences, respectively.  In Step 3, we compute the cosine similarity between all pairs of citances and judgment sentences and place them in matrix $\mathcal S_{m \times n}$.   Thus element $\mathit {s}_{pq}$  of matrix  $\mathcal S$ denotes the semantic  similarity between the $p^{th}$ citance and $q^{th}$ judgment sentence. Next, we employ a sentence scoring method to identify the judgment sentences that are semantically close to the citances and garner the most attention among the citing judgments. The scoring method selects the significant sentences and returns the summary of desired length $\mathit {l}$. Selected sentences are rearranged according to the judgment's original order to ensure coherence. 

The sentence scoring function (Step 4) is the critical component of an unsupervised extractive summarization algorithm and primarily determines the summary quality. Our scoring approach relies on semantic similarity between the contextual information contained in the \textit{citing-text-spans} and the judgment sentences.  We describe three scoring heuristics and compare them empirically in Sec. \ref{sec:Experimental Results}. 

\subsection{Sentence Scoring}
\label{subSec:Sentence Scoring}
In citation-based summarization of judgment,  the objective is to identify judgment sentences that closely align with most citances. It is noteworthy that column $q$ in matrix $\mathcal S$ reflects the semantic similarity of the $q^{th}$ judgment sentence with all citances. Similarly, row $p$ indicates the similarity of $p^{th}$ citance with all judgment sentences.  We propose three approaches described in the subsections below.
\subsubsection{ CiSumm Sentence Scoring}
\label{sec-algo: CiSumm-Sentence-Scoring}
CiSumm scoring method is a simple and intuitive scoring method that considers all citances equally relevant for creation of summary. The method is based on the intuition that a judgment sentence that exhibits higher overall similarity with all citances deserves to be included in summary. The score of the judgment sentence $\mathbf{j_q}$ is the sum of similarity scores with all citances (i.e. column ${\mathcal S_{*q}}$ ). Top-scoring judgment sentences are selected as candidates for the summary.  

CiSumm scoring scheme is attractive due to its simplicity and efficiency. However, since judgments often have long and complex sentences, longer sentences  benefit because of their ability to ensconce richer context. Furthermore, Lengthy judgment sentences sometimes  result in unintended  semantic similarity with many citances, which may promote them into summary.  We examine  the effect of length normalization in  mitigating  the bias due to sentence length in  Sec. \ref{sec:Experimental Results}. 
\subsubsection{Additive Sentence Scoring} 
\label{sec-algo:Additive-Sentence-Scoring}
 Additive sentence scoring method follows the idea that importance of a judgment sentence is proportional to the number of citances with which it bears similarity. Accordingly, we construct a candidate list $\mathcal C$ of top\mbox{-}$k$ scoring judgment sentences for each citance.  Genuinely important sentences in the judgment are expected to be cited more often and hence may be repeated in the list. Higher number of  repetitions indicates that the judgment sentence is semantically similar to more citances, and hence must be important. Similarity scores of the repeated sentences are added to signify their exalted importance.  Thus, the additive scoring  approach elevates the scores of  the candidate sentences in proportion to the references they gather in citing judgments. These sentences are the prime candidates for inclusion in the summary. 

 \begin{algorithm}
\scriptsize
    \KwIn{Similarity Score matrix ${\mathcal S}$, desired summary length $\mathit {l}$} 
    \KwOut{Judgment Summary $JSumm$}
       For each citation sentence $s_i \in \mathcal S$, add top\mbox{-}$k$ scoring judgment sentences along with scores to candidate list $\mathcal C$ \\
       Sum up the scores of each repeated judgment sentence in $\mathcal C$ \\
       Select top-scoring sentences from $\mathcal C$ to create summary $JSumm$  of length $\mathit {l}$;\\
       Return $JSumm$
      \caption{Additive Sentence Scoring}
\label{algo:additive-Sentence-Scoring}
\end{algorithm}

Algorithm \ref{algo:additive-Sentence-Scoring} outlines the proposed additive scoring method.  In Step 1, each row (${\mathcal S_{p*}}$) of the matrix ${\mathcal S}$ is scanned corresponding to the citance ($\mathbf{s_p}$) and top\mbox{-}$k$ judgment sentences ($\mathbf{j_q}$'s) along with their respective scores ($s_{pq}$'s) are added to the candidate list $\mathcal C$.  Step 2 sums up the scores of the repeated judgment sentences, and finally, top-scoring sentences from $\mathcal C$ are selected to craft the summary $JSumm$  of desired length $\mathit{l}$. 

\subsubsection{Citation Diversity Sentence Scoring} 
\label{sec-algo:Citation-Diversity-Sentence-Scoring}
 In \textit{C}itation \textit{D}iversity (CD) sentence scoring method,  we select judgment sentences while considering the diverse context in which the judgment is cited. This is accomplished by ensuring that each citance (i.e., each row in matrix $\mathcal S$) is processed and the corresponding high-scoring judgment sentence is given due consideration. 

We select top-$k$ judgment sentences along with their scores for each citance ($s_i$) and length normalized them. The post-selection length normalization avoids inclusion of shorter judgment sentences, which may not penalized sufficiently if length normalization is done before selection.

For each citation sentence, we arrange judgment sentences in descending order of their length-normalized score. At this juncture, the relevance of the citances is considered by scrutinizing the similarity scores of the top-scoring judgment sentences for all citances. The judgment sentence with the highest score is the one that has the strongest semantic similarity among all citances. The sentence is picked up and added to the candidate set along with the score. Proceeding with a scoring judgment sentence for the next strong citance, the candidate set is created. Finally, top-scoring sentences are selected to construct the summary $JSumm$.

\begin{algorithm}
	\scriptsize
        \KwIn{Similarity Score matrix ${\mathcal S}$, desired summary length $\mathit{l}$ } %
 	\KwOut{Judgment Summary $JSumm$  }
        For each citation sentence $s_i \in \mathcal S$, add top\mbox{-}$k$ scoring judgment sentences along with scores to list $L_i$ \\
        Normalize similarity score of each judgment sentence in $L_i$ by its length \\
        From each $L_i$, select judgment sentence most similar to $s_i$ and add to candidate set $\mathcal{C}$ if it's not already present \\
        Select the sentences from $\mathcal{C}$ in descending order of similarity scores to complete the summary $JSumm$\\
        If desired summary length is not completed, revise $\mathcal{C}$ by repeating Step 3 by considering the next most similar judgment sentence from each list $\{L_1, \ldots, L_m\}$ as in Step 3\\
        Repeat Steps 4 and 5, until the desired summary length $\mathit{l}$ is achieved \\
        Return $JSumm$\\

 \caption{Citation Diversity Sentence Scoring}
\label{algo-Citation Diversity Sentence Scoring}
\end{algorithm}

  Algorithm \ref{algo-Citation Diversity Sentence Scoring} describes the pseudo-code of the CD scoring method. Step 1 creates the list  $L_i$ of $k$ top-scoring judgement sentences along with their corresponding scores for each citance $s_i$. The scores are revised by normalizing them  by the respective  sentence length in Step 2. From the list $L_i$, the judgment sentence that closely matches the citance $s_i$ is included in the candidate set $\mathcal{C}$ if not added yet (Step 3) . This permits efficient comparison of the relevance and diversity of the citances based on their similarities with judgment sentences. Step 4 creates the summary $JSumm$ by choosing the sentences from $\mathcal{C}$ in descending order of scores. If the summary  falls shorter than desired summary length $\mathit{l}$, the next most resembling judgment sentence is added to $\mathcal{C}$ from the lists $L_i$'s (Step 5). Steps 4 and 5 are repeated until the desired summary length is completed (Step 6).


\section{Experiment and Results}
\label{sec:Experimental Results}

\subsection{Datasets}
\label{subSec:Datasets}
To assess the performance of  \textit{CB-JSumm} algorithm, we curate two datasets\footnote{\url{https://github.com/PurnimaBindal/LegalTextSummarization}} tailored for citation-based legal summarization in the Indian context. To the best of our knowledge, no existing dataset  leverages judgment citations for summarization of landmark legal judgments. 

The first dataset, \textit{IN-Jud-Cit}, consists of fifty landmark judgments from Indian Courts. We meticulously gathered citing judgments from the  \textit{Indian Kanoon}\footnote{ \url{https://indiankanoon.org/}} website using APIs provided with a free account. 
For each judgment, the gold-standard summary is obtained from \textit{Casemine}\footnote{\url{https://www.casemine.com/}} website. Interestingly, these summaries are AI-generated (abstractive) and shorter in length. 
Hence, we anticipate that the proposed scoring methods will yield a lower ROUGE score with small variation.

The second dataset, \textit{IN-Ext-Cit},  is curated by upgrading the IN-Ext dataset authored by \citet{Shukla2022} and comprising fifty Indian Court judgment-summary pairs.  Each judgment in IN-Ext dataset has two associated gold standard summaries written by two different law experts.  As before, we obtain citing judgments from the \textit{Indian Kanoon} website for each judgment in IN-Ext dataset.  One judgment, for which there was no citation, was discarded, and the remaining  49 judgments were used for experiments. Final score for each evaluation metric is obtained by averaging the score of two reference summaries.
\begin{table}[ht!]
\centering
\resizebox{\columnwidth}{!}{%
  \begin{tabular}{|c|c|c|c|c|c|c|}
    \hline
     & & &\multicolumn{2}{c|}{\textbf{Judgment Statistics}} & \multicolumn{2}{c|}{\textbf{Summary Statistics}}\\
    \hline
    \textbf{Dataset} & \textbf{$J$} & \textbf{$CJ$} & \textbf{$Sent$}  & \textbf{$W$} & \textbf{$Sent$} & \textbf{$W$}   \\
    \hline
    
    \textit{IN-Jud-Cit} & 50 & 15 &  259  &  8915  &  20 &  465  \\
    \hline
    \textit{IN-Ext-Cit} & 49 & 14 &  109 &  3775  &  47 &  1375  \\
    \hline
    \end{tabular}
    }
    \caption{Statistics of two datasets. J: judgments in the dataset, CJ:  average number of citing judgments, Sent: median number of sentences,  W: median number of words in the judgments}
    \label{Datasets-used-in-this-work-table}
\end{table}


\begin{table*}[ht!]
\centering
\resizebox{\textwidth}{!}{%
\begin{tabular}{|c|l|lccccc|}
\hline
\multirow{3}{*}{\textbf{Algorithm}} &
  \multicolumn{1}{c|}{\multirow{3}{*}{\textbf{Scoring Methods}}} &
  \multicolumn{3}{c|}{\textbf{\textit{IN-Jud-Cit} Dataset }} &
  \multicolumn{3}{c|}{\textbf{\textit{IN-Ext-Cit}} Dataset} \\ \cline{3-8} 
 &
  \multicolumn{1}{c|}{} &
  \multicolumn{6}{c|}{\textbf{ROUGE-F Scores}} \\ \cline{3-8} 
 &
  \multicolumn{1}{c|}{} &
  \multicolumn{1}{c|}{\textbf{Rouge-1}} &
  \multicolumn{1}{c|}{\textbf{Rouge-2}} &
  \multicolumn{1}{c|}{\textbf{Rouge-L}} &
  \multicolumn{1}{c|}{\textbf{Rouge-1}} &
  \multicolumn{1}{c|}{\textbf{Rouge-2}} &
  \multicolumn{1}{c|}{\textbf{Rouge-L}} \\ \hline
\multirow{5}{*}{\textbf{\textit{CB-JSumm}}} &
 CiSumm ($\S$ \ref{sec-algo: CiSumm-Sentence-Scoring})&
  \multicolumn{1}{c|}{$\mathbf{52.13  \pm 6.46}$} &
  \multicolumn{1}{c|}{$19.10 \pm 6.44$} &
  \multicolumn{1}{c|}{$54.48 \pm 6.39$} &
  \multicolumn{1}{c|}{$64.60 \pm 5.91$} &
  \multicolumn{1}{c|}{$38.62 \pm 9.34$} &
  66.74 ± 5.68 \\ \cline{2-8} 
  &
  CiSumm-LN ($\S$ \ref{sec-algo: CiSumm-Sentence-Scoring})&
  \multicolumn{1}{c|}{$49.40 \pm 7.46$} &
  \multicolumn{1}{c|}{$18.06 \pm 7.63$} &
  \multicolumn{1}{c|}{$53.79 \pm 6.88$} &
  \multicolumn{1}{c|}{$65.05 \pm 6.13$} &
  \multicolumn{1}{c|}{$40.22 \pm 9.91$} &
  67.95 ± 5.66 \\ \cline{2-8} 
 &
  Additive ($\S$\ref{sec-algo:Additive-Sentence-Scoring}) &
  \multicolumn{1}{c|}{$50.83 \pm 5.99$} &
  \multicolumn{1}{c|}{$17.93 \pm 5.14$} &
  \multicolumn{1}{c|}{$53.39 \pm 5.69$} &
  \multicolumn{1}{c|}{$64.15 \pm 6.38$} &
  \multicolumn{1}{c|}{$38.00 \pm 9.47$} &
  $66.37 \pm 6.07$ \\ \cline{2-8} 
 &
  Additive-LN ($\S$\ref{sec-algo:Additive-Sentence-Scoring}) &
  \multicolumn{1}{c|}{$50.27 \pm 7.35$} &
  \multicolumn{1}{c|}{$18.07 \pm 7.19$} &
  \multicolumn{1}{c|}{$53.30 \pm 6.90$} &
  \multicolumn{1}{c|}{$64.85 \pm 5.85$} &
  \multicolumn{1}{c|}{$39.17 \pm 9.09$} &
  $67.25 \pm 5.44$ \\ \cline{2-8} 
 &
  CD ($\S$\ref{sec-algo:Citation-Diversity-Sentence-Scoring})&
  \multicolumn{1}{c|}{$51.65 \pm 6.20$} &
  \multicolumn{1}{c|}{$\mathbf{19.90 \pm 7.12}$} &
  \multicolumn{1}{c|}{$\mathbf{55.61 \pm 5.71}$} &
  \multicolumn{1}{c|}{$\mathbf{65.49 \pm 5.98}$} &
  \multicolumn{1}{c|}{$\mathbf{40.29 \pm 9.38}$} &
  $\mathbf{68.04 \pm 5.70}$ \\ \hline
\multirow{4}{*}{\textbf{Competing}} &
  CaseSummarizer &
  \multicolumn{1}{c|}{$40.17 \pm 7.43$} &
  \multicolumn{1}{c|}{$10.16 \pm 6.25$} &
  \multicolumn{1}{c|}{$44.11 \pm 6.86$} &
  \multicolumn{1}{c|}{$59.44 \pm 6.28$} &
  \multicolumn{1}{c|}{$31.64 \pm 8.58$} &
  $61.94 \pm 6.00$ \\ \cline{2-8} 
 &
  MMR &
  \multicolumn{1}{c|}{$51.48 \pm 8.93$} &
  \multicolumn{1}{c|}{$19.79 \pm 8.58$} &
  \multicolumn{1}{c|}{$54.72 \pm 8.44$} &
  \multicolumn{1}{c|}{$57.80 \pm 6.30$} &
  \multicolumn{1}{c|}{$27.97 \pm 8.44$} &
  $60.38 \pm 6.04$ \\ \cline{2-8} 
 &
  Legal-pegasus &
  \multicolumn{1}{c|}{$47.76 \pm 13.63$} &
  \multicolumn{1}{c|}{$18.28 \pm 8.79$} &
  \multicolumn{1}{c|}{$50.41 \pm 13.21$} &
  \multicolumn{1}{c|}{$61.97 \pm 5.84$} &
  \multicolumn{1}{c|}{$32.77 \pm 8.12$} &
  $64.20 \pm 5.61$ \\ \cline{2-8} 
 &
  Legal-LED &
  \multicolumn{1}{c|}{$37.89 \pm 6.01$} &
  \multicolumn{1}{c|}{$10.15 \pm 3.47$} &
  \multicolumn{1}{c|}{$42.36 \pm 5.50$} &
  \multicolumn{1}{c|}{$49.03 \pm 5.42$} &
  \multicolumn{1}{c|}{$22.96 \pm 5.89$} &
  $52.55 \pm 5.05$ \\ \hline
\end{tabular}%
}
\caption{Macro-averaged ROUGE F-scores along with std. deviation for the two datasets. LN: Length-normalization.}
\label{tab:Result-Rouge}
\end{table*}

Table \ref{Datasets-used-in-this-work-table} summarizes the statistics for judgments, summaries, and citing judgments for both datasets. We  report median statistics for words and sentences due to substantial variability in judgment lengths within the datasets.   It is observed that judgments in \textit{IN-Jud-Cit} dataset are lengthier than those in \textit{IN-Ext-Cit}.  The reference summaries in \textit{IN-Jud-Cit} dataset are approximately $5\%$ of the original judgment length, whereas \textit{IN-Ext-Cit} summaries are approximately $36\%$ of the judgment length.
 \subsection{Competing Methods and Evaluation Metrics}
We assess \textit{CB-JSumm's}  performance for three sentence scoring methods (with and without sentence length normalization), with four competing legal-domain algorithms: CaseSummarizer\citep{polsley2016casesummarizer}, MMR\citep{Shukla2022}, Legal-pegasus\cite{nsi319/legal-pegasus} and Legal-LED\cite{nsi319_legal_led_base_16384}, obtained from GitHub repository\footnote{\url{https://github.com/Law-AI/summarization}}. 

We report macro-averaged ROUGE F-scores (ROUGE-1, ROUGE-2 and ROUGE-L). We augment our investigation by adopting the semantic-based assessment metric introduced by \citet{steinberger2009evaluation}. 


\subsection{Experimental Results}
We report experimental results for two datasets separately, as macro-averaged values along with the standard deviations.  

\noindent\textit{Results for \textit{IN-Jud-Cit} Dataset}:
Table \ref{tab:Result-Rouge} exhibits \textit{CB-JSumm's} superior performance over four competing methods. All scoring methods surpass CaseSummarizer, which is corpus-dependent and uses \textit{tf-idf} for sentence scoring. ROUGE scores for CD sentence scoring method are competitive with MMR scores, but other scoring methods show marginal performance decline. MMR algorithm also employs \textit{tf-idf} for sentence scoring and achieves diversity by repetitively scoring judgment sentences, which slows down the algorithm. Legal-pegasus and Legal-LED's degraded performance aligns with \citet{Shukla2022} results.

As evident from Table \ref{tab:Result-SemanticSimilarity}, the citation-diversity scoring method outperforms other scoring methods and  competing algorithms. MMR and Legal-Pegasus perform better than additive-scoring methods, while the performances of CaseSummarizer and Legal-LED leave much to be desired. 
\begin{table}[ht!]
\centering
\resizebox{\columnwidth}{!}{%
\begin{tabular}{|c|l|c|c|}
\hline
\textbf{Algorithm} & \multicolumn{1}{c|}{\textbf{Scoring Methods}} & \textbf{\textit{IN-Jud-Cit}} & \textbf{\textit{IN-Ext-Cit}  } \\ \hline
\multirow{5}{*}{\textbf{\textit{CB-JSumm}}} & CiSumm ($\S$ \ref{sec-algo: CiSumm-Sentence-Scoring}) & $0.75 \pm 0.12$           & $\mathbf{0.93 \pm 0.05}$          \\ \cline{2-4} 
                                            & CiSumm-LN ($\S$ \ref{sec-algo: CiSumm-Sentence-Scoring}) & $0.76 \pm 0.11$          & $0.92 \pm 0.04$          \\ \cline{2-4} 
                                            & Additive ($\S$ \ref{sec-algo:Additive-Sentence-Scoring})               & $0.74 \pm 0.11$          & $\mathbf{0.93 \pm 0.04}$ \\ \cline{2-4} 
                                            & Additive-LN  ($\S$ \ref{sec-algo:Additive-Sentence-Scoring})  & $0.74 \pm 0.12$          & $\mathbf{0.93 \pm 0.04}$ \\ \cline{2-4} 
                                            
                                            & CD ($\S$ \ref{sec-algo:Citation-Diversity-Sentence-Scoring}) & $\mathbf{0.78 \pm 0.09}$  & $\mathbf{0.93 \pm 0.05}$          \\ \hline
\multirow{4}{*}{\textbf{Competing}}         & CaseSummarizer            & $0.44 \pm 0.18$          & $0.90 \pm 0.08$          \\ \cline{2-4} 
                                            & MMR                       & $0.76 \pm 0.16$          & $0.92 \pm 0.05$           \\ \cline{2-4} 
                                            & Legal-pegasus             & $0.76 \pm 0.19$          & $0.92 \pm 0.05$          \\ \cline{2-4} 
                                            & Legal-LED                 & $0.60 \pm 0.13$          & $0.65 \pm 0.13$          \\ \hline
\end{tabular}%
}
\caption{Macro-averaged semantic similarity scores with standard deviation between system and reference summary for both datasets. LN: Length-normalization.}
\label{tab:Result-SemanticSimilarity}
\end{table}

\noindent\textit{Results for \textit{IN-Ext-Cit} Dataset}:
ROUGE scores are comparatively higher for this dataset owing to longer summaries (Table \ref{tab:Result-Rouge}). \textit{CB-JSumm} consistently outperforms competing algorithms with a bigger margin due to similar number of citing judgments and considerably shorter judgment length than \textit{IN-Jud-Cit} dataset (Table \ref{Datasets-used-in-this-work-table}).   This not only vindicates the importance of citations for summarization of landmark judgments, it also explains the narrow winning margin for \textit{IN-Jud-Cit} dataset, where the judgments are much longer but adequate citations are not available for summarization. Further,  CD scoring method slightly outperforms other scoring methods across all ROUGE variations.


Semantic similarity scores for this dataset are also higher due to lengthy  summaries for shorter judgments. As evident in Table \ref{tab:Result-SemanticSimilarity}, \textit{CB-JSumm} algorithm performs better than other competing methods.  Legal-LED and CaseSummarizer under perform, while MMR and Legal-pegasus slightly lag behind the proposed algorithm.


As stated before, \textit{IN-Jud-Cit} dataset reference summaries are short and abstractive, limit ROUGE score diversity among the proposed sentence scoring methods. For \textit{In-Ext-Cit} dataset, CD scoring performs best among all, as the reference summary contains sentences of the original judgment. Higher scores of this method indicate that post-selection length-normalization excludes very short judgment sentences favored by pre-selection normalization.



\section{Conclusion}
We propose \textit{CB-JSumm}, an extractive and unsupervised algorithm to summarize landmark judgments leveraging contextual information from citing judgments.  We evaluate proposed algorithm using two curated dataset and observe encouraging results.


\section*{Acknowledgements}

We thank authors \citet{Shukla2022} for sharing IN-Ext dataset. We also extend our sincere appreciation to Vineet Kumar and Irfan Sheikh for their dedicated efforts in developing the dataset pipeline.

\newpage

\bibliography{custom}

\begin{thebibliography}{22}
\expandafter\ifx\csname natexlab\endcsname\relax\def\natexlab#1{#1}\fi

\bibitem[{Anand and Wagh(2022)}]{anand2022effective}
Deepa Anand and Rupali Wagh. 2022.
\newblock Effective deep learning approaches for summarization of legal texts.
\newblock \emph{Journal of King Saud University-Computer and Information Sciences}, 34(5):2141--2150.

\bibitem[{Bhattacharya et~al.(2019)Bhattacharya, Hiware, Rajgaria, Pochhi, Ghosh, and Ghosh}]{bhattacharya2019comparative}
Paheli Bhattacharya, Kaustubh Hiware, Subham Rajgaria, Nilay Pochhi, Kripabandhu Ghosh, and Saptarshi Ghosh. 2019.
\newblock A comparative study of summarization algorithms applied to legal case judgments.
\newblock In \emph{Advances in Information Retrieval: 41st European Conference on IR Research, ECIR 2019, Cologne, Germany, April 14--18, 2019, Proceedings, Part I 41}, pages 413--428. Springer.

\bibitem[{Bhattacharya et~al.(2021)Bhattacharya, Poddar, Rudra, Ghosh, and Ghosh}]{bhattacharya2021incorporating}
Paheli Bhattacharya, Soham Poddar, Koustav Rudra, Kripabandhu Ghosh, and Saptarshi Ghosh. 2021.
\newblock Incorporating domain knowledge for extractive summarization of legal case documents.
\newblock In \emph{Proceedings of the eighteenth international conference on artificial intelligence and law}, pages 22--31.

\bibitem[{Chalkidis et~al.(2021)Chalkidis, Jana, Hartung, Bommarito, Androutsopoulos, Katz, and Aletras}]{chalkidis2021lexglue}
Ilias Chalkidis, Abhik Jana, Dirk Hartung, Michael Bommarito, Ion Androutsopoulos, Daniel~Martin Katz, and Nikolaos Aletras. 2021.
\newblock Lexglue: A benchmark dataset for legal language understanding in english.
\newblock \emph{arXiv preprint arXiv:2110.00976}.

\bibitem[{Farzindar(2004)}]{farzindar2004atefeh}
Atefeh Farzindar. 2004.
\newblock Atefeh farzindar and guy lapalme,'letsum, an automatic legal text summarizing system in t. gordon (ed.), legal knowledge and information systems. jurix 2004: The seventeenth annual conference. amsterdam: Ios press, 2004, pp. 11-18.
\newblock In \emph{Legal knowledge and information systems: JURIX 2004, the seventeenth annual conference}, volume 120, page~11. IOS Press.

\bibitem[{Feijo and Moreira(2023)}]{feijo2023improving}
Diego de~Vargas Feijo and Viviane~P Moreira. 2023.
\newblock Improving abstractive summarization of legal rulings through textual entailment.
\newblock \emph{Artificial intelligence and law}, 31(1):91--113.

\bibitem[{Furniturewala et~al.(2021)Furniturewala, Jain, Kumari, and Sharma}]{furniturewala2021legal}
Shaz Furniturewala, Racchit Jain, Vijay Kumari, and Yashvardhan Sharma. 2021.
\newblock Legal text classification and summarization using transformers and joint text features.

\bibitem[{Galgani et~al.(2012{\natexlab{a}})Galgani, Compton, and Hoffmann}]{galgani2012citation}
Filippo Galgani, Paul Compton, and Achim Hoffmann. 2012{\natexlab{a}}.
\newblock Citation based summarisation of legal texts.
\newblock In \emph{PRICAI 2012: Trends in Artificial Intelligence: 12th Pacific Rim International Conference on Artificial Intelligence, Kuching, Malaysia, September 3-7, 2012. Proceedings 12}, pages 40--52. Springer.

\bibitem[{Galgani et~al.(2012{\natexlab{b}})Galgani, Compton, and Hoffmann}]{galgani2012combining}
Filippo Galgani, Paul Compton, and Achim Hoffmann. 2012{\natexlab{b}}.
\newblock Combining different summarization techniques for legal text.
\newblock In \emph{Proceedings of the workshop on innovative hybrid approaches to the processing of textual data}, pages 115--123.

\bibitem[{Ghosh et~al.(2022)Ghosh, Dutta, and Das}]{ghosh2022indian}
Satyajit Ghosh, Mousumi Dutta, and Tanaya Das. 2022.
\newblock Indian legal text summarization: A text normalization-based approach.
\newblock In \emph{2022 IEEE 19th India Council International Conference (INDICON)}, pages 1--4. IEEE.

\bibitem[{Jain et~al.(2021)Jain, Borah, and Biswas}]{jain2021summarization}
Deepali Jain, Malaya~Dutta Borah, and Anupam Biswas. 2021.
\newblock Summarization of indian legal judgement documents via ensembling of contextual embedding based mlp models.

\bibitem[{Kalamkar et~al.(2022)Kalamkar, Agarwal, Tiwari, Gupta, Karn, and Raghavan}]{kalamkar2022named}
Prathamesh Kalamkar, Astha Agarwal, Aman Tiwari, Smita Gupta, Saurabh Karn, and Vivek Raghavan. 2022.
\newblock Named entity recognition in indian court judgments.
\newblock \emph{arXiv preprint arXiv:2211.03442}.

\bibitem[{Liu and Chen(2019)}]{liu2019extracting}
Chao-Lin Liu and Kuan-Chun Chen. 2019.
\newblock Extracting the gist of chinese judgments of the supreme court.
\newblock In \emph{proceedings of the seventeenth international conference on artificial intelligence and law}, pages 73--82.

\bibitem[{NSI319(2021{\natexlab{a}})}]{nsi319_legal_led_base_16384}
NSI319. 2021{\natexlab{a}}.
\newblock Legal language model (led) base 16384.
\newblock \url{https://huggingface.co/nsi319/legal-led-base-16384}.

\bibitem[{NSI319(2021{\natexlab{b}})}]{nsi319/legal-pegasus}
NSI319. 2021{\natexlab{b}}.
\newblock Legal pegasus.
\newblock \url{https://huggingface.co/nsi319/legal-pegasus}.

\bibitem[{Parikh et~al.(2021)Parikh, Mathur, Mehta, Mittal, and Majumder}]{parikh2021lawsum}
Vedant Parikh, Vidit Mathur, Parth Mehta, Namita Mittal, and Prasenjit Majumder. 2021.
\newblock Lawsum: A weakly supervised approach for indian legal document summarization.
\newblock \emph{arXiv preprint arXiv:2110.01188}.

\bibitem[{Paul et~al.(2022)Paul, Mandal, Goyal, and Ghosh}]{paul2022pre}
Shounak Paul, Arpan Mandal, Pawan Goyal, and Saptarshi Ghosh. 2022.
\newblock Pre-training transformers on indian legal text.
\newblock \emph{arXiv preprint arXiv:2209.06049}.

\bibitem[{Polsley et~al.(2016)Polsley, Jhunjhunwala, and Huang}]{polsley2016casesummarizer}
Seth Polsley, Pooja Jhunjhunwala, and Ruihong Huang. 2016.
\newblock Casesummarizer: A system for automated summarization of legal texts.
\newblock In \emph{Proceedings of COLING 2016, the 26th international conference on Computational Linguistics: System Demonstrations}, pages 258--262.

\bibitem[{Saravanan et~al.(2006)Saravanan, Ravindran, and Raman}]{saravanan2006improving}
Murali Saravanan, Balaraman Ravindran, and Shivani Raman. 2006.
\newblock Improving legal document summarization using graphical models.
\newblock \emph{Frontiers in Artificial Intelligence and Applications}, 152:51.

\bibitem[{Shukla et~al.(2022)Shukla, Bhattacharya, Poddar, Mukherjee, Ghosh, Goyal, and Ghosh}]{Shukla2022}
Abhay Shukla, Paheli Bhattacharya, Soham Poddar, Rajdeep Mukherjee, Kripabandhu Ghosh, Pawan Goyal, and Saptarshi Ghosh. 2022.
\newblock Legal case document summarization: Extractive and abstractive methods and their evaluation.
\newblock In \emph{The 2nd Conference of the Asia-Pacific Chapter of the Association for Computational Linguistics and the 12th International Joint Conference on Natural Language Processing}.

\bibitem[{Steinberger and Je{\v{z}}ek(2009)}]{steinberger2009evaluation}
Josef Steinberger and Karel Je{\v{z}}ek. 2009.
\newblock Evaluation measures for text summarization.
\newblock \emph{Computing and Informatics}, 28(2):251--275.

\bibitem[{Zheng et~al.(2021)Zheng, Guha, Anderson, Henderson, and Ho}]{zheng2021does}
Lucia Zheng, Neel Guha, Brandon~R Anderson, Peter Henderson, and Daniel~E Ho. 2021.
\newblock When does pretraining help? assessing self-supervised learning for law and the casehold dataset of 53,000+ legal holdings.
\newblock In \emph{Proceedings of the eighteenth international conference on artificial intelligence and law}, pages 159--168.

\end{thebibliography}

\end{document}